\documentclass{bmvc2k}
\usepackage{color}
\usepackage[accsupp]{axessibility}  

\usepackage{microtype}
\usepackage{adjustbox}
\usepackage{listings}
\usepackage{pythonhighlight}
\usepackage{float}
\usepackage{tablefootnote}
\usepackage{tabularx}
\usepackage{booktabs}
\usepackage{graphicx}
\usepackage{float}
\usepackage{wrapfig}

\usepackage{graphicx}
\usepackage{amsmath}
\usepackage{amssymb}
\usepackage{booktabs}

\usepackage[capitalize]{cleveref}
\crefname{section}{Sec.}{Secs.}
\Crefname{section}{Section}{Sections}
\Crefname{table}{Table}{Tables}
\crefname{table}{Tab.}{Tabs.}

\def\etal{\emph{et al}\bmvaOneDot}

\newcommand{\DECIG}{$\mathsf{DECIG}$}
\DeclareSymbolFont{Xlargesymbols}{OMX}{cmex}{m}{n}
\DeclareMathSymbol{\Xsum}{\mathop}{Xlargesymbols}{80}

\title{Towards Device Efficient Conditional Image Generation}
\addauthor{Nisarg A. Shah}{snisarg812@gmail.com}{1}
\addauthor{Gaurav Bharaj}{first.last@gmail.com}{1}

\addinstitution{
 AI Foundation\\
 California, USA
}

\runninghead{Shah and Bharaj}{DECIG: Device Efficient Conditional Image Generation}

\begin{document}

\maketitle

\begin{abstract}
We present a novel algorithm to reduce tensor compute required by a conditional image generation autoencoder without sacrificing quality of photo-realistic image generation. Our method is device agnostic, and can optimize an autoencoder for a given CPU-only, GPU compute device(s) in about normal time it takes to train an autoencoder on a generic workstation. We achieve this via a two-stage novel strategy where, first, we condense the channel weights, such that, as few as possible channels are used. Then, we prune the nearly zeroed out weight activations, and fine-tune the autoencoder. To maintain image quality, fine-tuning is done via student-teacher training, where we reuse the condensed autoencoder as the teacher. We show performance gains for various conditional image generation tasks: segmentation mask to face images, face images to cartoonization, and finally CycleGAN-based model over multiple compute devices. We perform various ablation studies to justify the claims and design choices, and achieve real-time versions of various autoencoders on CPU-only devices while maintaining image quality, thus enabling at-scale deployment of such autoencoders.
\end{abstract}

\section{Introduction}
\label{sec:introduction}
High demand for consumer avatars, filters, and scene generation applications has led to an increased at-scale need of photo-realistic image generation. Such applications rely on Generative Adversarial Networks (GANs)~\cite{goodfellow2014generative} and supervised image-to-image style transfer~\cite{gatys2016image,isola2017image,wang2018video} via autoencoders such as U-nets~\cite{ronneberger2015u}. Technical advancements, and availability of deep learning APIs~\cite{paszke2019pytorch,abadi2016tensorflow} has helped achieve image generation. Backends of such APIs rely on fast tensor operations, parallelized via GPUs. However, real-time image generation via GAN-like methods has high deployment cost due to GPU compute costs and high break-even profitability point. Although certain edge devices are native GPUs~\cite{wang2020practical} capable, they can suffer from slow inference, quality and resolutions deterioration of generated images. Thus, we desire a solution that can quickly optimize neural-nets for a given compute device, without sacrificing image quality and inference times. State-of-the-art literature suggests several approaches -- neural architecture design (NAD)~\cite{howard2017mobilenets,iandola2016squeezenet}, network architecture search (NAS)~\cite{zoph2018learning}, neural-net compression (quantization~\cite{han2015deep}, distillation~\cite{polino2018model}, and pruning~\cite{he2017channel,han2015deep}). We propose a novel net pruning algorithm that be employed along with NAD and NAS.
\\
\begin{figure*}[t]
\centering
  \includegraphics[width=0.95\linewidth]{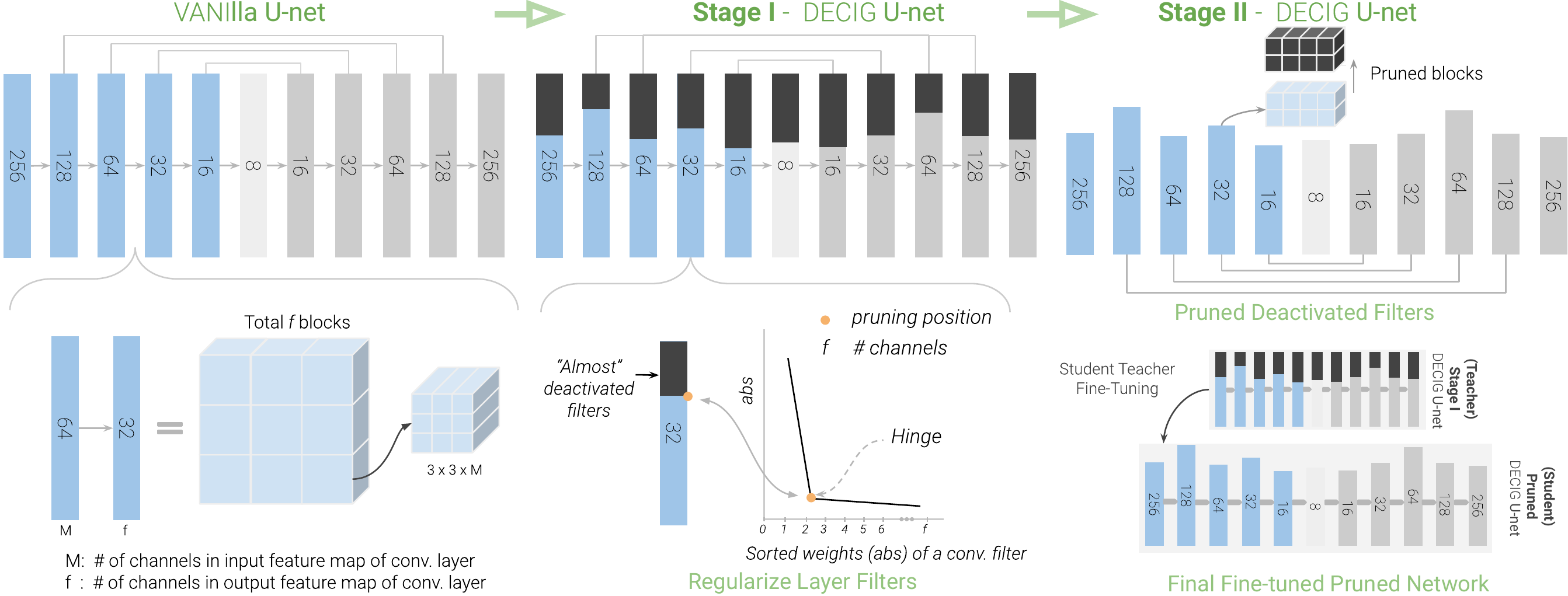}
  \caption{Our method \DECIG, dynamically condenses channel filter and prunes GAN-based autoencoders for image generation. During Stage I, vanilla U-net autoencoder is trained with penalization, where weight distribution (centre-bottom) has several resulting near-zero value channels that can be pruned. During Stage II, the pruned network is fine-tuned in student-teacher manner using condensed model (from Stage-I) as the teacher.}
  \label{fig:system-overview}
\end{figure*}
Typical neural-net model compression techniques focus on image classification and detection~\cite{he2019filter}. Applying such methods for conditional image generation is relatively less explored and a na\"ive application may lead to image artifacts. Shu~\etal~\cite{shu2019co} propose a channel pruning GAN compression evolutionary search algorithm, however, their method is designed for cyclic-consistency image generation~\cite{zhu2017unpaired}, and nontrivial to extend for non-cyclic consistency GANs~\cite{isola2017image,sanakoyeu2018style}. Shu~\etal~\cite{shu2019co} show that generators compressed by classifier compression methods \cite{liu2017learning} suffer performance decay compared to original generators. Chen~\etal~\cite{chen2020distilling} propose GAN compression by training efficient generators by model distillation and remove dependency on cyclic consistency. Their student network is handcrafted and requires significant architectural engineering for good performance.
\\
We propose a novel strategy to create compute efficient autoencoders for a given device. We do this by condensing neural-net channel filter weight distribution that condense filter usage, and later, prunes least activated filters while fine-tune using student-teacher model, where, the condensed autoencoder acts as the teacher. Our method is device agnostic, and optimizes neural-nets for given device. This also allows for a trade-off between computation complexity, and synthesized image quality, Fig~\ref{fig:results_unet}. We summarize our novel contributions below:

\begin{enumerate}
\item A novel strategy to reduce compute costs via dynamic channel filter condensing and pruning GAN autoencoders for image generation.
\item A filter penalization loss for better filter weight distribution for easy pruning across layers, and detection of a ``hinge'' to get a minimum threshold for a particular filter structure, obtaining compute efficient autoencoders.
\item Optimized autoencoders perform real-time inference on CPU-only, CPU-GPU compute, with equivalent FID vs. vanilla autoencoders for conditional image generation.
\end{enumerate}

\section{Related Works}
\label{sec:relatedworks}
\paragraph{Conditional Image Generation} Goodfellow~\etal~\cite{goodfellow2014generative}'s tremendous success with image generation led to several vision and graphics applications \cite{ledig2017photo,park2019semantic,tewari2020stylerig}. For image-to-image tasks, such as, segmentation masks to image generation~\cite{isola2017image,wang2018high,park2019semantic}  aim to achieve better image quality, these methods require high-end (Nvidia's~\href{https://www.nvidia.com/en-us/data-center/v100}{V100}) GPUs for fast inference. At-scale use of such methods requires lower compute, and inference costs. Reducing neural-net compute is an active area of research. While most tackle non-image generation tasks, we explicitly aim to solve for conditional image generation. Such methods can be categorized as (1) Neural Architecture Design, (2) Neural Architecture Search, and (3) Neural-net Compression. Our work falls under the last category; we reduce compute by designing an algorithm that given an input autoencoder efficiently prunes filters, while maintaining generated image quality.
\paragraph{Neural Architecture Design} of CNNs~\cite{krizhevsky2017imagenet,he2016deep,simonyan2014very} led to massive gains for vision tasks. Such deep architectures are heavy on compute, even more so as autoencoders~\cite{ronneberger2015u}. While several lightweight architectures have been proposed~\cite{iandola2016squeezenet,hu2018squeeze,howard2017mobilenets,sandler2018mobilenetv2,zhang2018shufflenet,tan2019efficientnet,han2020ghostnet}, these works exploit costly tensor blocks within architectures, and replace them with lightweight ones, or perform tensor compute in an efficient manner to improve the performance, generally for non-image generation. Wang~\etal~\cite{wang2020practical} use depth-wise separable convolutions to reduce tensor compute for denoising. On the other hand, our method is a novel strategy to reduce compute given an autoencoder, and can leverage these architectures as input.
\paragraph{Neural Architecture Search} \textit{algorithmically} search for efficient architectures. This search is highly nonlinear, with high compute and time complexity. Several works search architectures via reinforcement learning, and genetic algorithms~\cite{stanley2002evolving,miikkulainen2019evolving,brock2017smash}. Zoph~\etal~\cite{zoph2018learning} search for transferable network blocks, surpasses manually designed architectures~\cite{szegedy2015going,he2016identity}. While Cai~\etal~\cite{cai2017reinforcement} speed up exploration for better architectures via network transformation. Fu~\etal~\cite{fu2020autogan}'s distiller framework does adaptive search for operators types and channel widths. AutoML~\cite{li2020gan} framework searches for channel widths for existing generators which can be computationally expensive. In comparison, our method optimizes accuracy and weight distribution using a penalization loss and trains in a similar time as vanilla autoencoder. Also, our approach is complementary and can be combined with NAS.
\paragraph{Classification based pruning methods for conditional Image generation}
Though, use of classification(\textit{Encoder only}) pruning methods seem to be a logical first choice for pruning in conditional image generation, several SOTA methods note otherwise. Yu et al.~\cite{yu2020self} conduct extensive experiments using \textit{standard} pruning methods -- manual pruning~\cite{han2015learning} and gradual pruning (AGP)~\cite{zhu2017prune} for GAN-based tasks (face generation). They note that image generated from manual pruning is of very poor quality, whereas, the models didn't converge using AGP~\cite{zhu2017prune}. \cite{wang2020gan} and \cite{li2021revisiting} highlight the instability in GAN minimax training and not being able to achieve Nash's equilibrium for retaining performance metrics (FID), on applying classification based pruning methods for conditional image generation. Similarly, Shu et al.\cite{shu2019co} show that applying classification neural-net compression \cite{liu2017learning, luo2017thinet,hu2016network} for generators compression suffers performance decay compared with the original generator. Based on these arguments we conclude that a na\"ive application of classification neural-net compression for image generation is less suitable, and thus the need for our approach.

\paragraph{Neural-Net Compression (Distillation, Quantization, Pruning)} Wang~\etal~\cite{wang2020gan}'s unified GAN compression framework uses model distillation, channel pruning, and quantization to address this issue. Their channel pruning method is based on \cite{liu2017learning}'s batch-norm~\cite{ioffe2015batch} scale parameters, and measures importance among kernels of a convolutional layer. This dependence on batch-norm makes it inherently infeasible for architectures without batch-norm, while we seek a more generic approach. Aguinaldo~\etal~\cite{aguinaldo2019compressing} compress DC-GAN~\cite{radford2015unsupervised} via knowledge distillation. Hou~\etal~\cite{hou2020slimmable} explore a method to generate nets with different channel sizes, while it leads to identity preserving image generation for face images, their FIDs may deteriorate. Lin~\etal~\cite{lin2021anycost} propose a student-teacher learning method for interactive photo-realistic image generation and Lin~\etal~\cite{li2021revisiting} propose a online collaborative distillation scheme to learn intermediate features of teacher generator and discriminator to optimize the student generator. Compared, we don't have to pre-design student network and can condense a model using magnitude distribution plots and ``hinge'' modeling, Sec~\ref{sec:method}. Further, our algorithm can be used along with student-teacher training methods for joint training and device specific optimizations.

\section{Method}
\label{sec:method}
\begin{figure*}[t]
    \centering
  \includegraphics[width=\linewidth]{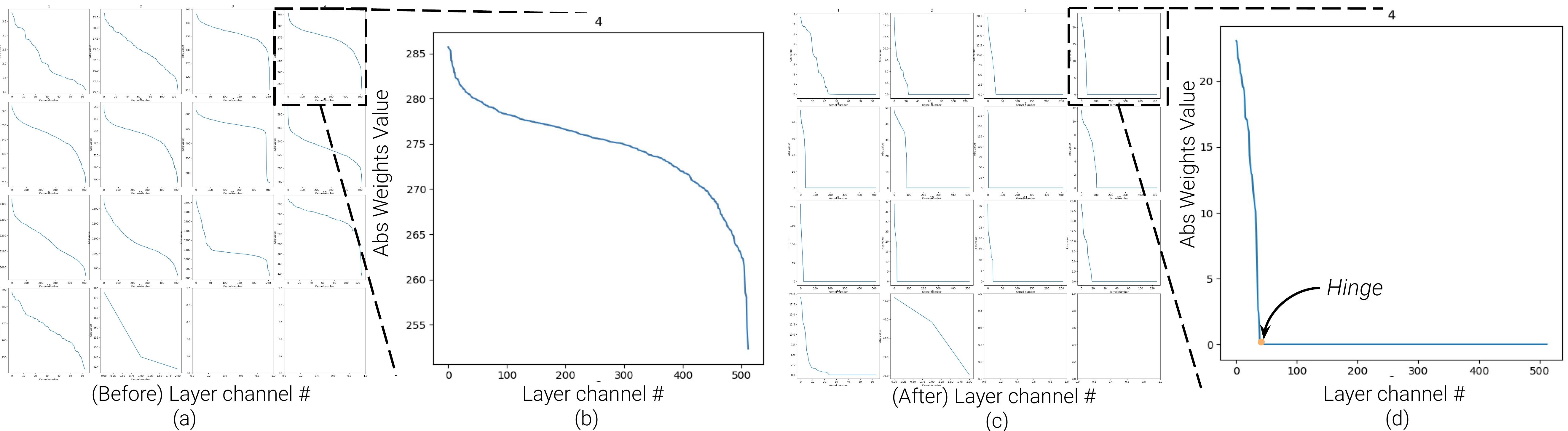}
  \caption{(a) \& (c) shows weight distribution of Unet-64's layers after training, without and with \DECIG~penalization; (b) and (d) shows zoomed-in version of corresponding \textit{fourth} layer. (d) show the ``hinge'' for pruning \textit{fourth} layer achieved by our method. X-axis indicates channel \# in the layer, Y-axis indicates absolute value of weights of corresponding channel.}
  \label{fig:small_layer_vis}
\end{figure*}
A vanilla autoencoder generator $\mathsf{G}$ learns to synthesize an image $\mathsf{I}$ from an input segmentation map, $S \in \{H\times W\times 3\}$. Our pix2pix-like~\cite{isola2017image} setup uses a U-net~\cite{ronneberger2015u} backbone generator, $\mathsf{G}$, while the optimized generator $\mathsf{G}^{*}$ aims to be compute efficient s.t. the quality of generated images from both generators ($\mathsf{G}, \mathsf{G}^*$) is nearly equivalent, while $\mathsf{G}^{*}$ can be deployed across diverse hardware -- CPUs, (e)GPUs optimizing for latency and image quality trade-off. The optimization condense (regularizes) filters used convolution layers of the autoencoder (Stage I), and later prunes least used filters (Stage II) and fine-tunes the pruned generator, Fig. \ref{fig:system-overview}.

As per, Li~\etal~\cite{li2016pruning} and Wen~\etal~\cite{wen2016learning} of the three levels for sparsity regularization, \emph{coarse} channel-level sparsity provides a better trade-off between flexibility and ease of deployment. It can be applied on any neural-net with convolutional layers, and generates a \emph{sparser} version of the original model. Channel-level sparsity requires pruning all adjacent connections associated with a particular channel, and makes it challenging to apply it directly on a pre-trained model due to generally non-existent zero weight channels in the neural-net, see Fig.~\ref{fig:small_layer_vis} (a). To alleviate the problem of non-existent zero weight channels for sparsity regularization, we enforce a penalization loss in the training objective. Specifically, we introduce a loss function that operates on absolute value of filter weights and systematically pushes the filter weights towards zero during training. 

Unlike Liu~\etal~\cite{liu2017learning} that regularize added scaling factors after convolution or on adjacent scaling factor of normalization layer, our method operates on layer weights. We observe that using extra scaling factors not only add computational burden, but also such normalization specific methods increase the complexity of the approach when dealing with new methods with pre-activation structures and cross-connecting layers like ResNets~\cite{he2016deep}, and DenseNets~\cite{huang2017densely}. Further, such methods designed with batch-norm (or normalization layers in general) become unusable when working with newer, normalization free architectures. Our loss function operates on magnitude of channel weights, and works with such newer architectures.\footnote{A layer indicates convolutional layer in the deep neural network, where a collection of channels make a single layer. For instance, if there is a $3\times3$ convolutional layer, with the input feature map of shape $H \times W \times M$ and output feature map of shape $H \times W \times N$, there would be total $N$ channels each with shape $3 \times 3 \times M$.}  

\textbf{Stage I - Channel Weight Regularization}
Channel pruning methods~\cite{lin2021anycost}, utilize kernel magnitude as the criterion for relative importance across filters. On the other hand, when we train a network, a per channel importance factor $\gamma$, equivalent to magnitude of the weights of the corresponding channels is introduced. We train the network weights, and optimize the importance factor to condense the weights to as few channels as possible. This training objective for the $i^{th}$ layer is given by:
\begin{align}
\mathsf{L}_{i} = \mathsf\Sigma_{j=1}^n \mathsf{f}(j) * ||\mathsf{W}_{i,j}||_1
\label{eq:reg}
\end{align}
Where, $n$ is layer \# in the network, $j$ the channel \# of the convolutional filter, and $\mathsf{W}_{i,j}$ the filter weight of the $i^{th}$ layer and $j^{th}$ sorted channel. We explore three different channel regularization strategies with $j \in (1,n)$:

\begin{enumerate}
    \item[-] \emph{Uniform} feature channel regularization, $\mathsf{f}(j) = 1.0$
    \item[-] \emph{Linear} feature channel regularization, $\mathsf{f}(j) = j$
    \item[-] \emph{Exponential} feature channel regularization, $\mathsf{f}(j) = e^{0.01j}$
\end{enumerate}
We have selected three different channel regularisation methods, i.e. linear, uniform and exponential, to explore and verify the effects of relative penalisation (faster increase in penalisation as number of channels increase for a layer) and its effects on compression and quality.

\begin{figure*}
\centering
  \includegraphics[width=\linewidth]{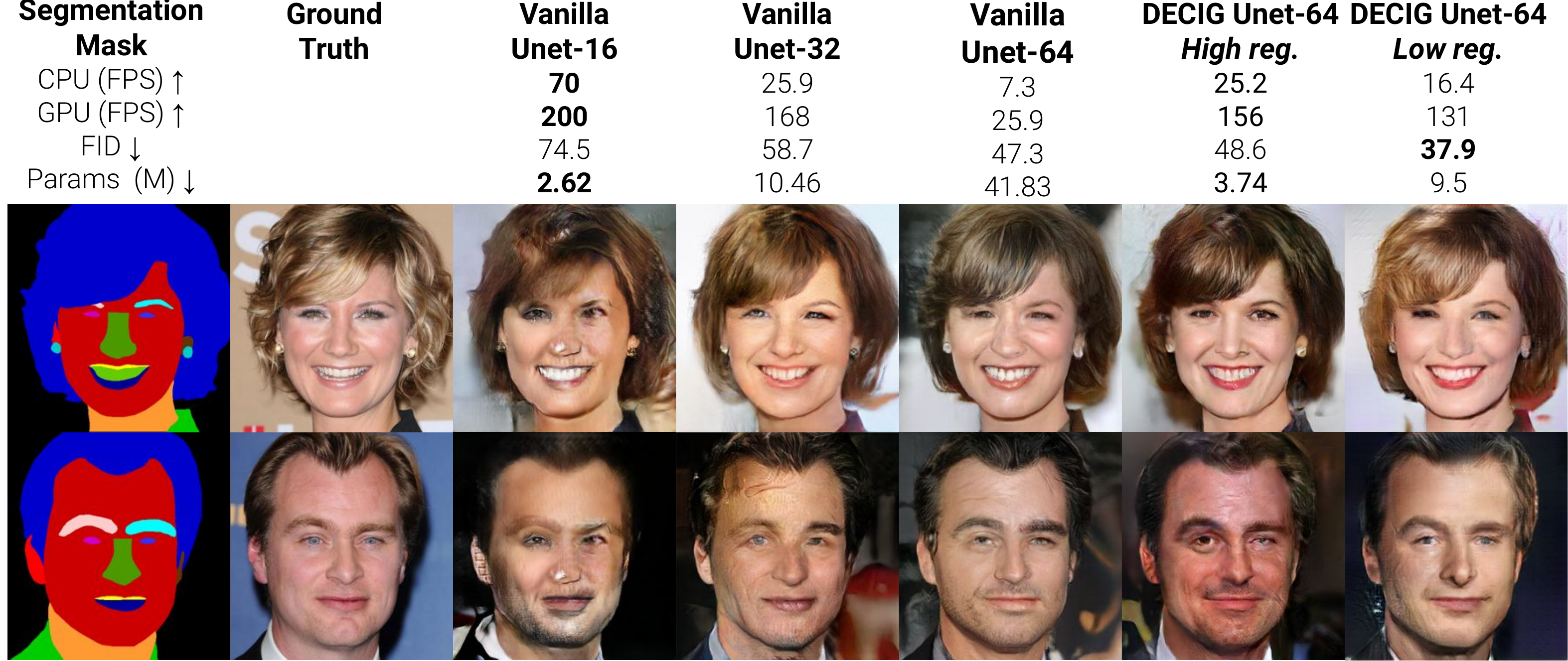}
  \caption{(Left to Right) Baseline variants: Unet-\{16, 32, 64\}, and \DECIG~versions of Unet-64 with high and low regularization. \textit{Note: While better image generation methods exist, our emphasis is to maintain image quality given baseline autoencoders.}}
  \label{fig:results_unet}
\end{figure*}

\textbf{Stage I - Layer Device Performance Regularization.} GPU devices exploit tensor compute parallelism in convolution layers and simultaneously process large number of weight channels. CPU devices, however, carry out these operations sequentially and don't benefit from similar parallelism. Depending on the device and their memory allocation, relative convolutional operation speed across different spatial resolutions and feature map sizes differs considerably. For example, a \texttt{convolution(kernel=3, stride=2)} at $8\times 8$ resolution with $512$ input and output channels require 7.179 milliseconds (ms) on CPU and 1.132 ms on GPU. However, same convolution at $16\times 16$ resolution takes 
$3\times$ time 
and 
$1.6\times$ time 
on CPU and GPU, respectively. Similarly, for a \texttt{convolution(kernel=3, stride=2)} at $128\times 128$ resolution with input channel $1$, if the number of output channels are increased from 32 to 128, the run-time for CPU gets quadrupled ($4\times$ cost) while for GPU it remains nearly same. Based on this insight, we make the neural-net optimization device specific.

For model deployment, the compute devices are usually fixed, thus we propose run-time layer level (device-dependent) channel regularization strategy. We calculate the run-time for each layer across a particular device, and use it as a multiplicative factor $\mathsf{l}(i)$ for that layer, to calculate total penalization. Our method allows device agnostic, multiply-accumulate (MAC) operations based, layer-level channel regularization. To this end, we calculate the multiplicative factor of each layer based on corresponding MAC operations of that particular layer. The penalization $\mathsf{L}_{\mathsf{PENAL}}$, minimax optimization~\cite{goodfellow2014generative} for a GAN is $\min_{\mathsf{G}} \max_{\mathsf{D}} \mathsf{L}_{\mathsf{GAN}}$ objective function, and final objective $\mathsf{L}_{\mathsf{GAN}}^{\mathsf{ALAP}}$ are given as:
\begin{align}
\mathsf{L}_{\mathsf{PENAL}} &= \mathsf\Sigma_{i=0}^{n} \mathsf{l}(i) * \mathsf{L}_{i}\nonumber\\
\mathsf{L}_{\mathsf{GAN}} &=\mathbb{E}_{y \in \mathcal{Y}}[\log(\mathsf{D}(y)]+\mathbb{E}_{x \in \mathcal{X}}[\log (1-\mathsf{D}(\mathsf{G}(x))]\nonumber\\
\mathsf{L}_{\mathsf{GAN}}^{\mathsf{ALAP}} &= \mathsf{L}_{\mathsf{GAN}} + \mathsf{L}_{\mathsf{l1}} +\alpha *\mathsf{L}_{\mathsf{PENAL}}
\label{eq:final_alap}
\end{align}
Where, $\mathcal{X}$ corresponds to random noise distribution, while $\mathcal{Y}$ corresponds to real image distribution. l1 loss (ground-truth and generated images) is used during training as well. $\alpha$ is decided such that, for the first training iteration, ratio of $\alpha *\mathsf{L}_{\mathsf{PENAL}}$ and $\mathsf{L}_{\mathsf{l1}}$ is 0.1

\textbf{Stage II - Pruning and Distillation} After Stage-I training, Eq~\ref{eq:final_alap}, we obtain a model with a considerable amount of inactivated (near zero weight) channels. Due to the penalization loss, distinction between near zero and important channels is easily identifiable, Fig.~\ref{fig:small_layer_vis}(d). The inclination point that shows the threshold between these two types of channels is identified as the ``hinge''. Fig~\ref{fig:small_layer_vis} shows an example resulting weight distribution plot. We sort channels of a layer in Unet-64 model by magnitude (importance factor), and the hinge is identified at $50^{th}$ channel for this particular layer. This way, we do not require to take an arbitrary guess or a global threshold~\cite{liu2017learning} on the number of channels to be pruned, making the trained model \textit{device efficient}. After identifying the ``hinge'', channels lower in magnitude than hinge channel are pruned. Along with pruning these channels, we also remove corresponding incoming and outgoing connections and weights across all layers and obtain a compact network with fewer -- parameters, run-time memory, and compute operations.

\begin{figure*}[t]
  \centering
  \includegraphics[width=\linewidth]{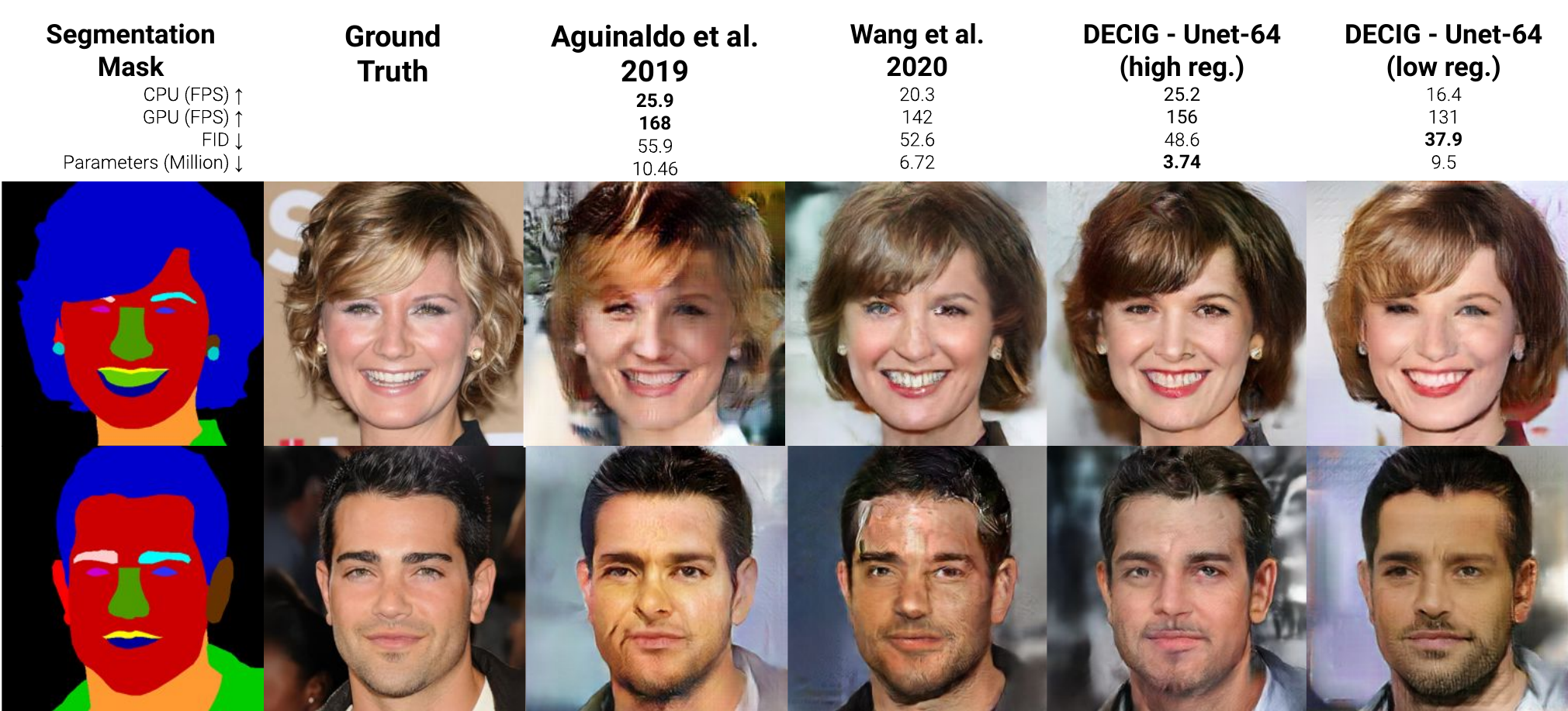}
  \caption{Comparison with state-of-art Distillation and
Pruning methods: (Left to Right) Segment Mask, G.T., Aguinaldo~\etal~\cite{aguinaldo2019compressing}, Wang~\etal~\cite{wang2020gan}, DECIG-Unet-64 (high reg.), DECIG-Unet-64 (low reg.)}
  \label{fig:ablation-sota}
\end{figure*}
This hinge-based pruning has a minimal effect on the perceptual quality of generated images because the pruned channels magnitude is far less compared to that of non-pruned ones, sharp slope in Fig.~\ref{fig:small_layer_vis}(d). We also observe that, \textit{almost} deactivated filters (Fig.~\ref{fig:system-overview}) act as noisy channels. Thus when pruned, the autoencoder generates images with higher perceptual quality without further training. The minimal negative effect on perceptual quality, if any, can be further nullified by fine-tuning the pruned network via student-teacher training, where Stage-I trained model acts as the teacher model. Additionally, in several cases, like the over-parameterized or low weight penalization models discussed in the subsequent section, we find that the fine-tuned pruned network can reach higher perceptual scores than the vanilla network. After this stage, we finally obtain the optimized generator $\mathsf{G}^*$.

\section{Results and Ablation}
\label{sec:results}
We present our results, comparison and ablation results for paired and unpaired conditional image generation tasks. Specifically, results with generator Unet-{16, 32, 64, 192} variants \footnote{Unet-x~\cite{isola2017image} indicates Unet autoencoder, where number of channels is $x$, that doubles after every strided convolution with an upper limit of $512$.} for pix2pix~\cite{isola2017image} and Resnet for CycleGAN for respective vanilla versions, and \DECIG~(ours) optimized models. Since, the discriminator does not affect inference time, the student and teacher discriminator structure was kept the same. 

To verify the generalizability across tasks and comparisons, we test \DECIG~on three applications: (1) segmentation mask to face generation, (2) face to cartoonized images for pix2pix model and (3) horse-to-zebra for CycleGAN-like model. We use CelebA-HQ dataset~\cite{CelebAMaskHQ}, that contains 30K high-quality face images and corresponding pixel-level segmentation masks face generation, and for cartoon images generation using animegan2\footnote{\url{https://github.com/bryandlee/animegan2-pytorch}}. We also evaluate the optimized generator via $\mathsf{G}^*$ using perceptual quality FID~\cite{heusel2017gans} score, memory consumption, run-times, and qualitative comparisons. We use MAC layer regularization and linear $\mathsf{f}(j)$, Eq.~\ref{eq:reg}, feature channel regularization for experiments, unless otherwise specified and measure the latency speed-up on Intel Xeon CPU E5-2686, and low-cost NVidia K80 GPU.

\begin{figure*}[t]
\centering
  \includegraphics[width=\linewidth]{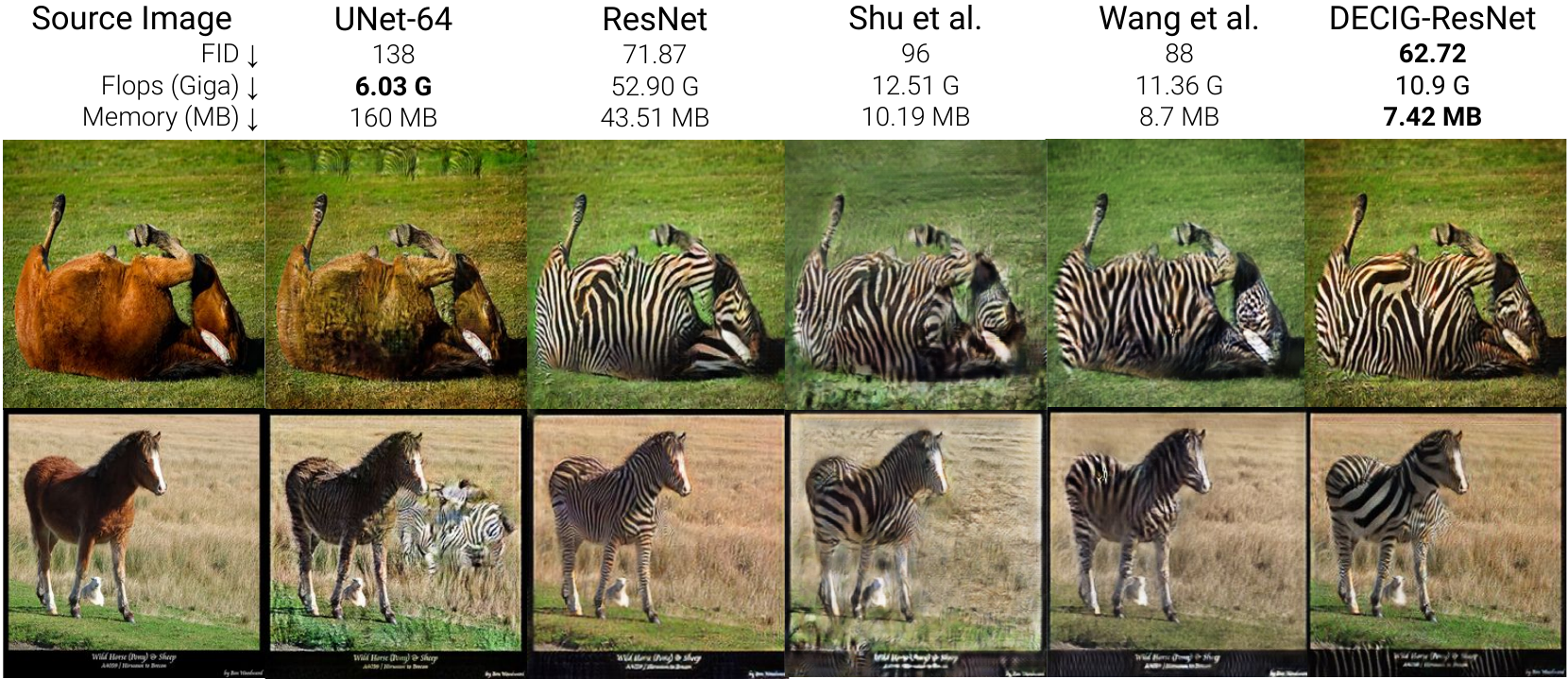}
  \caption{CycleGAN compression: (Left to Right) Source
Image, Style transfer UNet-64, ResNet CycleGAN, CEC \cite{shu2019co}, GAN-Sliming \cite{wang2020gan} and $\mathsf{DECIG}$-ResNet, respectively.} 
  \label{fig:h2z-images}
\end{figure*}
\textbf{Results \& Comparisons: Face Generation Application.} Fig.~\ref{fig:results_unet} shows results of variants on U-net~\cite{ronneberger2015u} architecture. Although, compared with Unet-64, \emph{miniature} Unet variants (Unet-16 and Unet-32)  have fewer 
parameters, their generated images have several artifacts -- austere and blurry repeated patches. Images generated by our condensed generators look sharper and more realistic with lower inference times. Our method (Fig.~\ref{fig:results_unet}) distributes weights in each layer, s.t., not only was it able to achieve better FID compared with corresponding vanilla models, but also hinge-based pruning had a minimal effect on FID. After student-teacher training, the model recovers from pruning artifacts and the resulting model has improved FIDs by 17\%, and run-time (FPS), parameters and memory improve significantly.

Inference time for $\mathsf{DECIG}$-Unet generators improves $3.5\times$ on CPU and $1.63\times$ on GPU. Primary objective of \textit{high-reg} version of $\mathsf{DECIG}$ is to enable higher model compression with equivalent perceptual scores, compared to it's vanilla variant, whereas, in \textit{low-reg} versions enables higher perceptual quality over compression metrics. \textit{high} and \textit{low} indicates the penalization in the overall loss function. \textit{high-reg} and \textit{low-reg} is a relative term, we have scaled high-reg with $0.1$ to get low-reg before training. The $\mathsf{DECIG}$-Unet-64 high-reg variant achieves a real-time inference on CPU-only device and makes it feasible for cost-effective deployment.

\textbf{Results \& Comparisons: CycleGAN application.} We go beyond Unets, and apply $\mathsf{DECIG}$ on unpaired image-to-to image translation, and use  ResNet~\cite{zhu2017unpaired} as our generator, which is more complex than a Unet due to its cross-connections, and compare with recent SOTA methods like GAN-sliming~\cite{li2020gan}. Following \cite{shu2019co,wang2020gan}, we use GFLops, model size (MB), and FID (between source style-transfer and target-style images) as a quantitatively measure for the ResNet generator.

On horse-to-zebra task, Fig.~\ref{fig:h2z-images}, \DECIG-ResNet achieves $5.9\times$ reduction in model size compared to baseline and a 30\% improvement in FID score. Our methods achieves significant improvement of 28\% in FID scores compared to Wang~\etal~\cite{wang2020gan} along with improved results over compression matrices. Using \textit{high-reg} version of \DECIG-ResNet we achieve 33\% reduction in the number of flops and $11\times$ reduction in memory usage with improvements in perceptual quality metrics, Table~\ref{fig:fig_sub_table}.

\begin{figure}[t]
\centering
  \includegraphics[width=\linewidth]{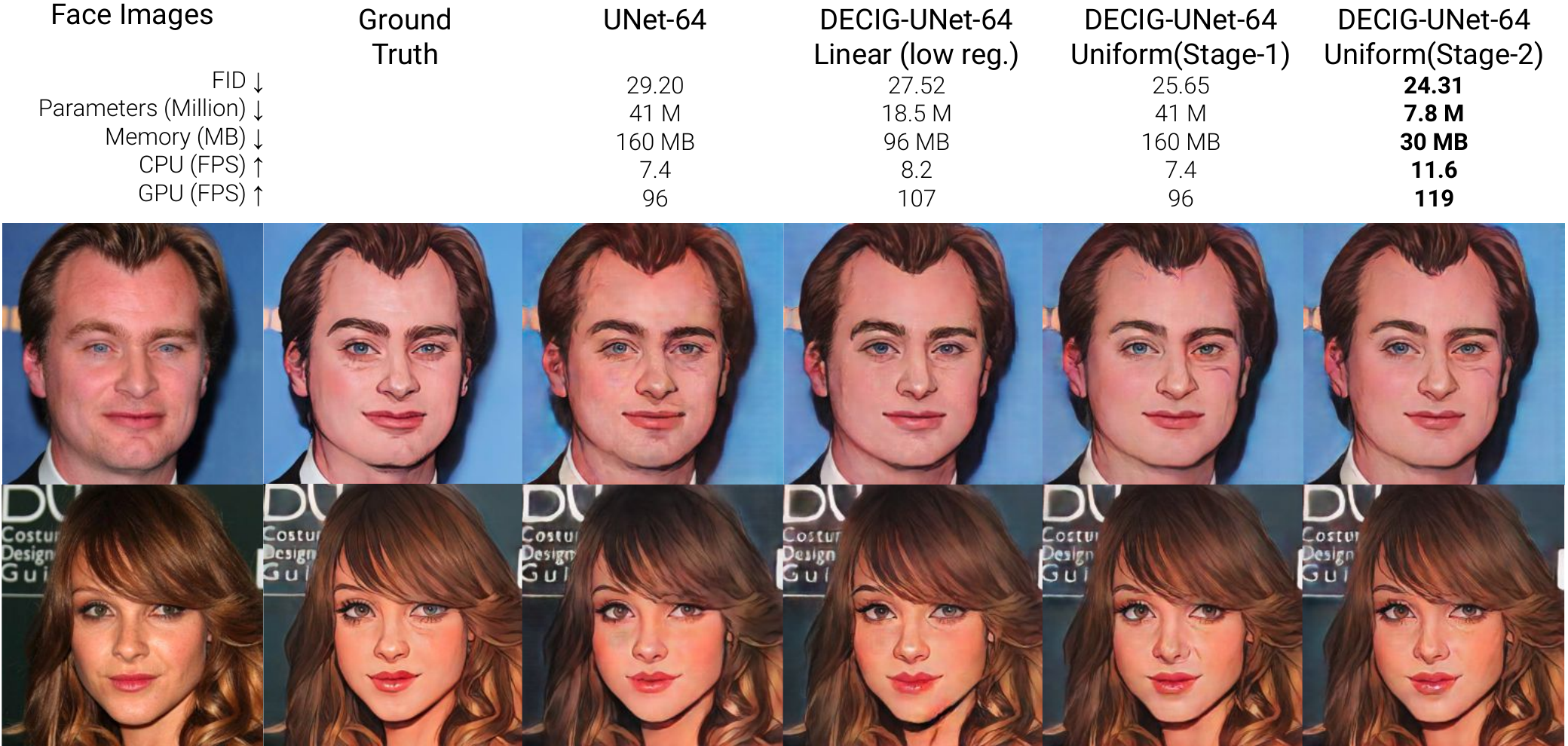}
  \caption{Face Cartoonization Compression: (Left to Right) Input, Cartoonized GT, UNet-64, $\mathsf{DECIG}$-UNet-64 Linear (low-reg), $\mathsf{DECIG}$-UNet-64 Uniform, Stage I and II.}
  \label{fig:cartoon-images}
\end{figure}

\textbf{Results: Cartoonization Application.}
Experiments are conducted on two different channel weight regularization approaches, i.e., Uniform and Linear, and the weighing of penalization is controlled to get balanced improvements over both FID scores and compression metrics.
The qualitative and quantitative results of the cartoon-style transfer, together with model statistics (FLOPs and model sizes), are shown in Fig. \ref{fig:cartoon-images}. We also show results after only stage-I training to quantify student-teacher's effect on our approach. Exploring improved student-teacher methods for training is orthogonal to our work and could be leveraged for additional improvements. For the cartoonization task, as fine-scale features are less critical, uniform channel weight regularization performs better compared with the linear version. $\mathsf{DECIG}$-UNet performs well compared to UNet, with a 20\% improvement in FID scores and 6$\times$ reduction in number of parameters. These results validate \DECIG's efficacy compared to models trained with the conventional approach in terms of perceptual quality and model parameter reduction.

\textbf{Ablation: Overparametrization and Regularization.}
We evaluate our method on \textit{overly-parametrized} Unet-192 and observe its advantages on over-fit reduction. \DECIG-Unet-192 (Fig.~\ref{fig:fig_sub_table}(b)) typically achieves a $6\times$ reduction in the number of parameters vs Unet-192. The pruned \DECIG-Unet-192 achieves a lower FID score (47.3, 30\% improvement) compared to the original model's FID (65.3). We attribute this to our penalization strategy that addresses the over-fitting over the training set; see supplementary for qualitative results.

\textbf{Ablation: Channel Weight and Layer Device Regularization} 
Our channel weight regularization method supports several multiplicative functions like Uniform, Linear, and Exponential, and the corresponding quantitative predictions are shown in Tab.~\ref{tbl:table_2}(b). 

In our experiments, we observe linear $\mathsf{f}(x)$ as the standard strategy based on trade-off perceptual image quality scores and run-time improvements, and uniform $\mathsf{f}(x)$ performs slightly better when fine-scale features are less critical
in the generation process like tasks such as cartoonization. We also test our approach on device agnostic layer level regularization (Tab.~\ref{tbl:table_2}(a)). Here, FPS for model's MAC optimized for CPU specifically (\DECIG-CPU) has better FPS on CPU device compared to model optimized for general device, see supplementary for qualitative results.

\textbf{Comparisons Summary with SOTA Pruning Methods.} Aguinaldo~\etal~\cite{aguinaldo2019compressing} uses model distillation to train a slim student generator from the corresponding bigger teacher network. We used Unet-64 as the teacher network and Unet-32, i.e., half slimmed channels for every layer as the student network and train using the distillation method. We observe slight FID improvement from vanilla Unet-32 model, however, generated results are blurry, Fig.~\ref{fig:ablation-sota}. We also implement Wang~\etal~\cite{wang2020gan} for vanilla Unet-64 and obtain improved results in terms of perceptual quality (FID), parameter reduction, and memory usage compared to~\cite{aguinaldo2019compressing}. However, images generated from \cite{wang2020gan} have several artifacts such as, random color patterns, illuminations and thus, quality of generated images have more artifacts that our methods, while both networks have similar sizes and runtime, Figure~\ref{fig:ablation-sota}.
\begin{table*}[]
\centering
\small
\begin{adjustbox}{width=\linewidth}
\centering
\begin{tabular}{cc|c|c|c||c|c|c|c}
\toprule
                               & Unet-64       & \begin{tabular}[c]{@{}c@{}}$\mathsf{DECIG}$\\ CPU\end{tabular} & \begin{tabular}[c]{@{}c@{}}$\mathsf{DECIG}$\\ GPU\end{tabular} & \begin{tabular}[c]{@{}c@{}}$\mathsf{DECIG}$\\ MAC\end{tabular} & \begin{tabular}[c]{@{}c@{}}Linear Pen. \\ (High reg.)\end{tabular} & \begin{tabular}[c]{@{}c@{}}Uniform \\ Penalization\end{tabular} & \begin{tabular}[c]{@{}c@{}}Exponential \\ Penalization\end{tabular} & \begin{tabular}[c]{@{}c@{}}Linear Pen. \\ (Low reg.)\end{tabular} \\ \midrule
\multicolumn{1}{l|}{CPU (FPS)} & 7.3           & \textbf{28.6}                                                  & 26.1                                                           & 25.2                                                           & \textbf{25.2}                                                      & 22.5                                                            & 21.9                                                                & 16.4                                                              \\ 
\multicolumn{1}{c|}{GPU (FPS)} & 96            & 161                                                            & \textbf{169}                                                   & 156                                                            & \textbf{156}                                                       & 149                                                             & 144                                                                 & 131                                                               \\ 
\multicolumn{1}{c|}{FID}       & \textbf{47.3} & 51.48                                                          & 51.61                                                          & 48.6                                                           & 48.6                                                               & 45.61                                                           & 44.46                                                               & \textbf{37.9}                                                     \\ \bottomrule

\end{tabular}
\end{adjustbox}
\caption{Comparison of Unet-64's \DECIG~variants with layer weight \{CPU, GPU, MAC\} and channel weight \{Linear (high-reg), Uniform, Exponential, Linear (low-reg)\} regularization.}
\label{tbl:table_2}
\end{table*}

\begin{figure}[t]
    \centering
    \begin{minipage}[t]{.47\textwidth}
\centering
\begin{adjustbox}{width=\linewidth}
\centering
\begin{tabular}{@{}r|c|c|c|cc@{}}
\toprule
\multicolumn{1}{l}{}               & ResNet & \begin{tabular}[c]{@{}c@{}} Shu et al. \\ \cite{shu2019co}\end{tabular} & \begin{tabular}[c]{@{}c@{}} Wang et al. \\ \cite{wang2020gan}\end{tabular} & \begin{tabular}[c]{@{}c@{}}$\mathsf{DECIG}$-\\ResNet \end{tabular} \\ \midrule
\multicolumn{1}{r|}{FID}     & 148.81 & 139.88     & 120         & \textbf{133}                                               \\
\multicolumn{1}{r|}{GFlops}     & 52.90  & 12.16      & 12.05       & \textbf{8.21}\\
\multicolumn{1}{r|}{Mem. (mb)}           & 43.51  & 10.02      & 9.04        & \textbf{4.83}      \\
\bottomrule
\end{tabular}
\end{adjustbox}
\par
(a)

    \end{minipage}%
    \hfill
    \begin{minipage}[t]{.48\textwidth}
\centering
\small
\begin{adjustbox}{width=\linewidth}
\centering
\begin{tabular}{l|l|l||l|l}
\toprule
           & \begin{tabular}[c]{@{}l@{}}Vanilla\\ ResNet- 128\end{tabular} & \begin{tabular}[c]{@{}l@{}}$\mathsf{DECIG}$ -\\ ResNet- 128\end{tabular} & \begin{tabular}[c]{@{}l@{}}Vanilla\\ Unet- 192\end{tabular} & \begin{tabular}[c]{@{}l@{}}$\mathsf{DECIG}$ -\\ Unet- 192\end{tabular} \\ \midrule
\textit{CPU, fps}  & 0.15                                                          & \textbf{0.63}                                                          & 0.45                                                        & \textbf{2.5}                                                         \\
\textit{GPU, fps}  & 3.43                                                          & \textbf{6.1}                                                           & 7.8                                                         & \textbf{25}                                                          \\
FID        & 52.2                                                          & \textbf{49.7}                                                          & 65.26                                                       & \textbf{42.8}                                                        \\
\textit{param} & 54                                                            & \textbf{4.9}                                                           & 376                                                         & \textbf{61.5}                                                        \\ \bottomrule
\end{tabular}
\end{adjustbox}
\par
(b)
    \end{minipage}
    \caption{(a) Comparison with CycleGAN Zebra-to-Horse Compression~\cite{shu2019co,wang2020gan} (b) Baseline conditional-GAN \{Unet-192, ResNet\}, and their $\mathsf{DECIG}$ versions.} 
\label{fig:fig_sub_table}
\end{figure}

\section{Limitations and Conclusion}
\label{sec:conclusion}
We present \DECIG, a tensor compute reduction method that optimizes autoencoders for conditional image generation and achieves near real-time inference capabilities on CPU-only, and GPU devices. Our method achieves significant improvements over state-of-the-art methods wrt run-time and perceptual quality for various conditional image generation tasks -- pix2pix (segmentation mask to images, images to cartoonization) and CycleGAN using autoencoders UNet and ResNet, respectively.
\\
Several limitations exist: Although, task specific, our method does not completely preserve face identity attribute after compression and requires manual penalization strategy. We also require picking the hinge during Stage-II pruning manually. In the future, we want to explore improvements for these limitations via better perceptual losses, and introduce identity preserving loss~\cite{deng2019arcface}, and propose automated hinge selection via clustering, and sharp curvature change modeling during compression aware training.

\bibliography{main}
\newpage
\maketitle
\section*{A. Detailed Setup and Implementation details}
We evaluate and ablated our method on CelebA-HQ dataset~\cite{CelebAMaskHQ} that contains 30,000 high-quality face images (resized to $256 \times 256$) and corresponding pixel-level segmentation mask annotations. We evaluate images generated by our optimized $\mathsf{G}^*$ using perceptual quality score -- FID ~\cite{heusel2017gans}, memory consumption, run-times, and quantitative comparisons. We conduct experiments to translate semantic segmentation mask to face images using pix2pix~\cite{isola2017image} to compare different methods. The paired dataset is divided into about 23,500 training images, 4,000 validation images, and about 2,500 test images. To verify the efficacy of our algorithm across different autoencoders, we follow the settings in pix2pix and use U-net~\cite{ronneberger2015u} and ResNet as generators. Like \cite{isola2017image,chen2020distilling}, we use PatchGANs, that uses $70\times 70$ image patches instead of whole images. During optimization of the networks, the objective value is divided by two while optimizing the discriminator. The networks are trained for $200$ epochs using Adam~\cite{kingma2014adam}, and learning rate of $1e^{-4}$. 

We use U-net~\cite{isola2017image} termed as Unet-64, where number of channels is $64$ and that gets doubled after every strided convolution with an upper limit of $512$. We also evaluate our approach on an \emph{overly-parametrized} Unet-192 to observe its advantages to reduce over-fitting. We also trained Unet-32 and Unet-16 to compare the pruned variants of Unet-64 in an equi-parametric setting. Since the discriminator does not affect inference time, the student and teacher discriminator structure was kept the same. We analyze the performance of different autoencoders -- Unet and ResNet, and compare their respective vanilla versions and optimized models using our proposed method. We further show application of $\mathsf{ALAP-AE}$ on CycleGAN for horse-to-zebra dataset, and on pix2pix to cartoonize the faces to verify the generalizability of our algorithm across different tasks and comparison with state-of-art methods available.
\section*{B. Detailed Qualitative and Quantitative results of \DECIG-UNet over CelebA-HQ dataset}

\begin{figure}
\centering
  \includegraphics[width=0.85\linewidth]{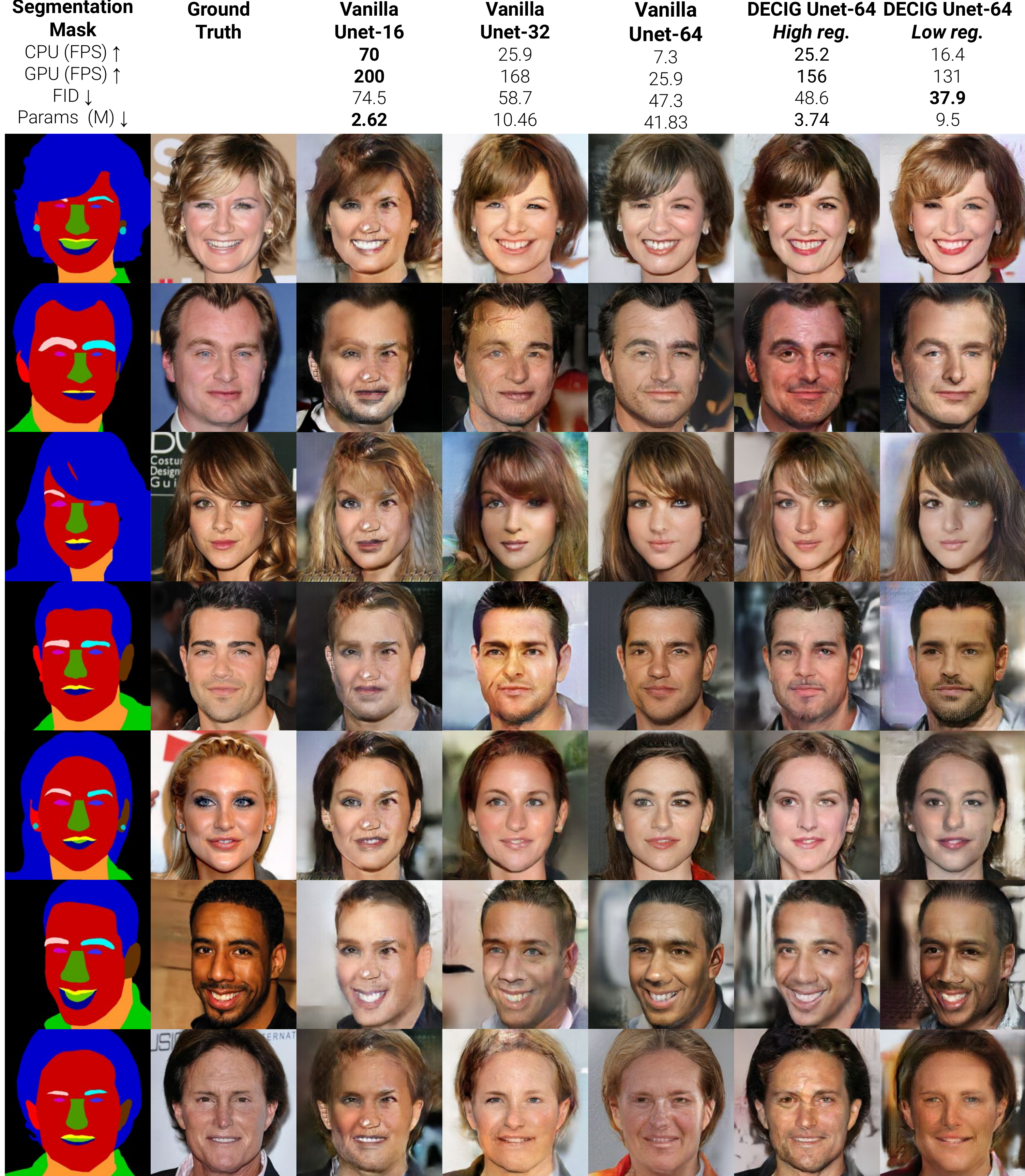}
  \caption{(Left to Right) We take several Vanilla Unet variants as baseline for conditional-GAN based Image generation (Unet-64, Unet-32, Unet-16), and create \DECIG versions of the Unet-64 network for high and low regularization setting. Note: While better image generation methods exist, our emphasis is to maintain image quality vs. the baselines autoencoders.}
  \label{fig:results_unet}
\end{figure}

 Fig.~\ref{fig:results_unet} shows additional results of several variants on U-net\cite{ronneberger2015u} architectures for conditional image generation. While satisfactory results are achieved for vanilla generator (Unet-64), it requires significant parameters as well as compute resources~\ref{fig:results_unet}. Although, \emph{miniature} Unet variants (Unet-16 and Unet-32) have fewer MACs (FLOPs), memory consumption, and parameters, their generated images look austere and blurry with repeated patches; thus making them look fake. While images generated by our proposed condensed generators look sharper and more realistic, at a low inference times. Here, it is important to note, that primary objective of \textit{high-reg} version of \DECIG~is to develop more compressed model with equivalent perceptual scores, compared to it's vanilla variant, whereas, in \textit{low-reg} versions, higher perceptual quality is preferred over compression metrics. \textit{High} and \textit{low} indicates the amount of penalization in the overall loss function. 

\section*{C. Channel Weight and Layer Device Regularization}
\begin{figure*}
  \centering
  \includegraphics[width=0.75\linewidth]{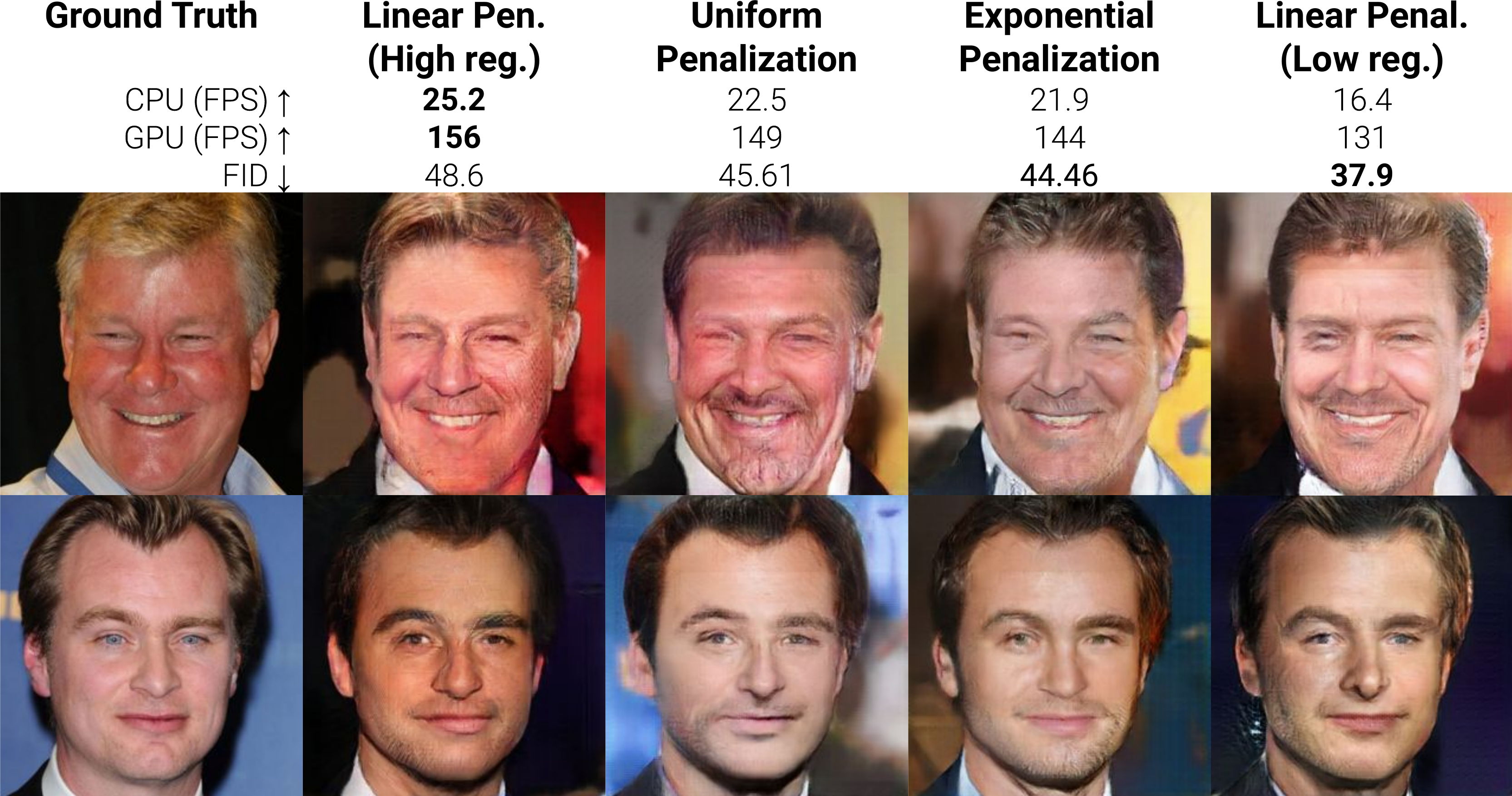}
  \caption{(Left to Right) We create the  versions of Unet-64 variant with different Channel weight regularization. ( \emph{Linear}(high-reg.), \emph{Uniform}, \emph{Exponential}, \emph{Linear}(low-reg.) feature channel regularization)}
  \label{fig:ablation-images}
\end{figure*}

Our channel weight regularization method supports several multiplicative functions like Uniform, Linear, and Exponential. Uniform and Exponential factors are used in low reg form. The qualitative and quantitative predictions for different channel weight multiplicative factors discussed  in Fig.~\ref{fig:ablation-images}.

\begin{figure}
  \centering
  \includegraphics[width=\linewidth]{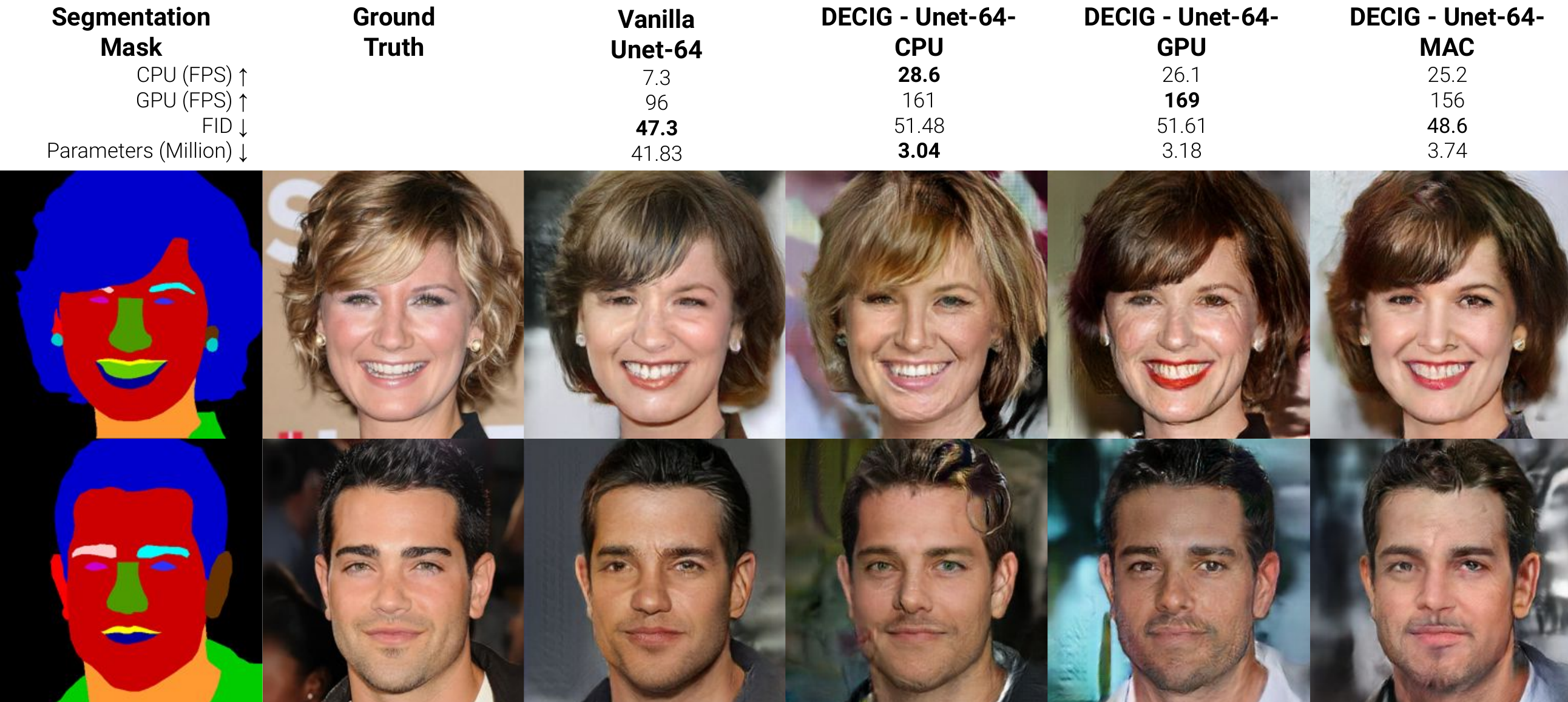}
  \caption{(Left to Right) Comparison of (high-reg) Unet-64 variants condensed for particular type of Device e.g. CPU, GPU and general (MAC) (Segmentation Map, Ground truth, Unet-64, $\mathsf{DECIG}$-Unet-64-[CPU, GPU, MAC]}
  \label{fig:ablation-cpu-gpu}
\end{figure}

We also tested the results for device agnostic layer level regularization discussed in Sec.~3.2 in Fig.~\ref{fig:ablation-cpu-gpu}. Here, specific improvements over inference time of model could be observed for the model optimised for that corresponding device i.e. CPU and GPU. 

\begin{figure}
  \centering
  \includegraphics[width=0.9\linewidth]{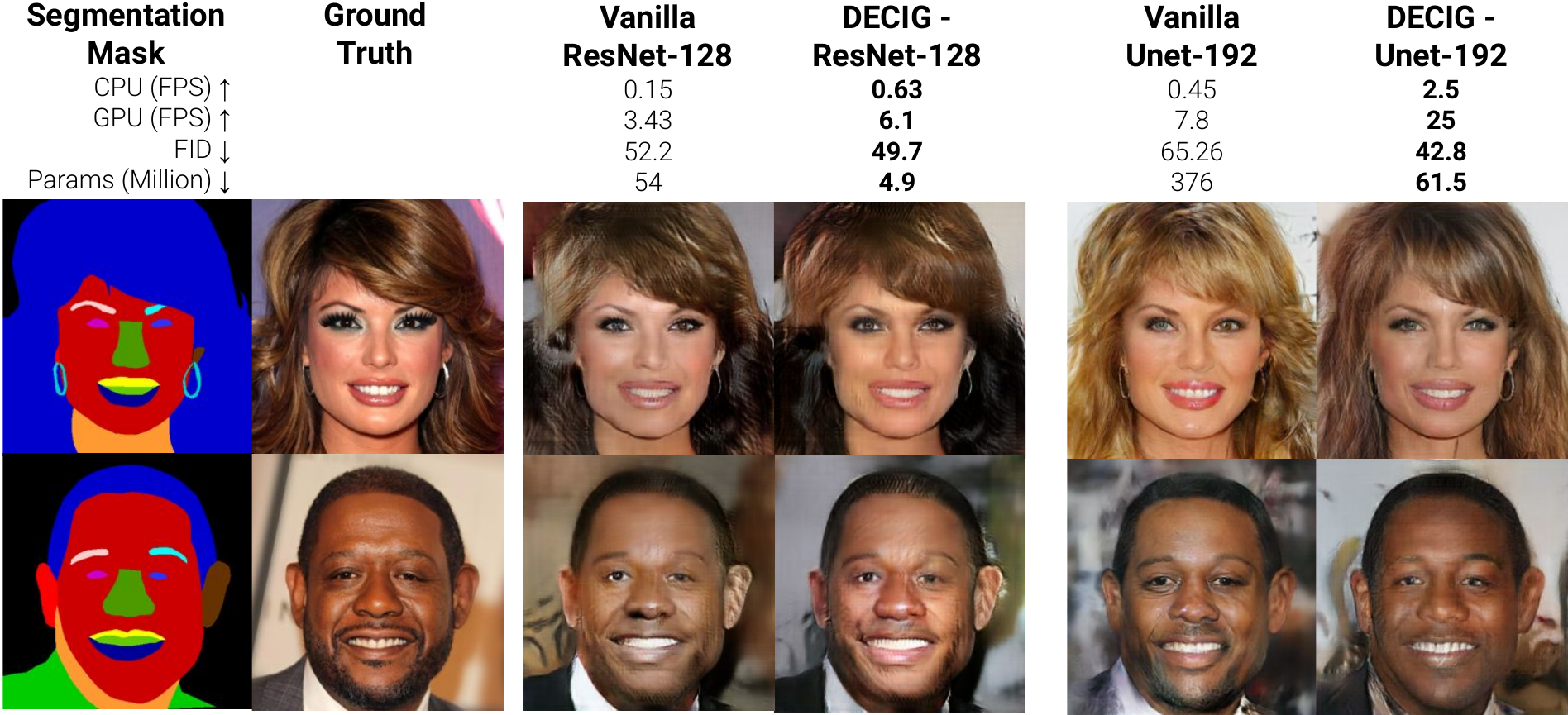}
  \caption{(Left to Right) We take baseline conditional-GAN based AE’s (Unet-192, ResNet), and create \DECIG ~versions of these AEs.}
  \label{fig:system}
\end{figure}

\section*{D. Overparametrization and Regularization effect}
On quantitative part of Fig.\ref{fig:system}, we can observe that, with Unet-192 and $\mathsf{DECIG}$-Unet-192, typically there's a $6\times$ reduction in the number of parameters, and the weight-induced pruned network achieves a lower FID score compared to the original model. Unet-192 has an FID score of $65.3$, which is 30\% poorer compared to its smaller variant Unet-64's FID of $47.3$. These 
results are produced due to over-fitting of the Unet-192 model on the training dataset. Interestingly, our penalization algorithm solves the over-fitting problem to an extent by achieving FID improvements of ~30\% with $5\times$ and $3.2\times$ improvements on run-time over CPU and GPU, respectively. We hypothesize this is due to the regularization effect of the penalization algorithm on channels that condenses the features in each layer, and make redundant channel weights zero.

\end{document}